\documentclass[manuscript]{acmart}

\AtBeginDocument{%
  \providecommand\BibTeX{{%
    \normalfont B\kern-0.5em{\scshape i\kern-0.25em b}\kern-0.8em\TeX}}}

\title{An Evaluation of LLMs for Detecting Harmful Computing Terms}

\author{Joshua Jacas}
\authornote{Both authors contributed equally to this research.}
\affiliation{%
  \institution{Clarksburg High School}
  \city{Clarksburg}
  \country{USA}
}

\author{Hana Winchester}
\authornotemark[1]
\affiliation{%
  \institution{The Ohio State University}
  \city{Columbus}
  \country{USA}
}

\author{Alicia Boyd}
\affiliation{%
  \institution{Yale University}
  \city{New Haven}
  \country{USA}
}

\author{Brittany Johnson}
\affiliation{%
  \institution{George Mason University}
  \city{}
  \country{USA}
}

\renewcommand{\shortauthors}{Jacas et al.}

\begin{document}

\begin{abstract}
    Detecting harmful and non-inclusive terminology in technical contexts is critical for fostering inclusive environments in computing. This study explores the impact of model architecture on harmful language detection by evaluating a curated database of technical terms, each paired with specific use cases. We tested a range of encoder, decoder, and encoder-decoder language models, including BERT-base-uncased, RoBERTa large-mnli, Gemini Flash 1.5 and 2.0, GPT-4, Claude AI Sonnet 3.5, T5-large, and BART-large-mnli. Each model was presented with a standardized prompt to identify harmful and non-inclusive language across 64 terms. Results reveal that decoder models, particularly Gemini Flash 2.0 and Claude AI, excel in nuanced contextual analysis, while encoder models like BERT exhibit strong pattern recognition but struggle with classification certainty. We discuss the implications of these findings for improving automated detection tools and highlight model-specific strengths and limitations in fostering inclusive communication in technical domains.
\end{abstract}



\maketitle

\section{Introduction \& Background}
\textbf{Content warning: This paper contains examples of stereotypes and associations,  erasure, and other harms that could be offensive and triggering to marginalized communities.}

In recent years, there has been a growing awareness of the importance of inclusive language within technical fields~\cite{giachanou2020battle, words2022acm, gilbert2022words,vanderborght2020united}. Non-inclusive terminologies, often ingrained in technical jargon, can perpetuate biases, reinforce stereotypes, and create spaces that are not welcoming to all individuals~\cite{winchester2023harmful}. This heightened awareness has spurred initiatives across industry and academia to identify and replace harmful terms with inclusive alternatives~\cite{words2022acm, winchester2022exploration, winchester2023hate}.

Inclusive language usage is crucial for fostering diverse and innovative environments. It helps dismantle barriers, enabling individuals from varied backgrounds to contribute fully and feel valued within their communities. 
As organizations strive to create more inclusive cultures while managing other costs and concerns, the need for automated tools that can detect non-inclusive language has become paramount.  

Although some harmful term detector tools such as Alex\footnote{
https://alexjs.com/} or Joblint\footnote{
https://joblint.org/} have been developed to address this need, they rely on rule-based methods. While these traditional techniques offer consistency and scalability, their rule-based nature can lead to inaccuracies and challenges in adapting to the nuanced use and evolution of language. 
Consequently, there is a growing interest in exploring the potential of leveraging artificial intelligence (AI) to enhance the detection of harmful and non-inclusive terminology~\cite{todd2024githubinclusifier}.
As the field of AI has advanced, large language models (LLMs) have emerged as promising solutions for text-based tasks. These models, with their ability to understand context and semantics at a deeper level, hold the potential to provide more accurate and context-sensitive assessments compared to traditional tools.

While the little work that has been done shows promise~\cite{todd2024githubinclusifier}, there is still much to learn about how to best leverage LLMs in this context. 
To this end, we conducted a study to investigate the impact of different LLM architectures on the ability to detect harmful and non-inclusive computing terminology.
Specifically, we investigate how various LLMs perform in identifying non-inclusive computing language when presented with the same prompt. 
The overarching research question guiding this study is: \textit{\textbf{How do different model architectures influence the ability to automatically detect harmful and non-inclusive computing terminology?}}

To answer this question, we conducted a comparative analysis of three LLM architectures: encoder, encoder-decoder, and decoder. 
Through this research, our objective is to advance the understanding of model-specific strengths and limitations in the context of inclusive language detection.
However, our analyses are situated within a larger research agenda aimed at supporting the detection, replacement, and explanation of harmful terms in computing.
 Ultimately, our goal is to contribute to developing robust and reliable tools that support ongoing efforts to create more inclusive technical environments. 






\subsection{Harmful Terminology in Technical Contexts.} 

Harmful or non-inclusive language refers to the use of terminology that can make others feel excluded or perpetuate stereotypes~\cite{winchester2023harmful}.
In computing, the most familar are terms such as ``master,'' ``slave,'' ``blacklist,'' and ``whitelist.''
As concerns around diversity, equity, and inclusion in computing continue to grow, we have seen widespread efforts (especially in industry) to replace harmful words like ``blacklist'' and ``whitelist'' with more inclusive terms like ``allowlist'' and ``denylist.''


\subsection{Automated Detection of Harmful Terms} 

Automated bias detection tools exist to support finding and fixing harmful language in writing. 
Existing tools analyze text for non-inclusive language and suggest alternatives or explain why changes are necessary.
For example, the HaTe Detector identifies harmful words and phrases with suggested edits~\cite{winchester2023hate}; the GitHub Inclusifier leverages large language models (LLMs) to identify non-inclusive terms in code and provide recommendations to make programming more inclusive~\cite{todd2024githubinclusifier}.
However, LLMs have been known to suffer from decreased accuracy when missing necessary or cultural context~\cite{varona2022discrimination}. 
Furthermore, models can have implicit human-like biases based on race, gender, and other social contexts~\cite{may-etal-2019-measuring, warr2024implicit, kumar2024investigating}.



\subsection{Impact of Model Architecture on NLP Tasks} 
Design and architecture considerations in AI models for natural language processeing (NLP) influences the capabilities of models in various scenarios. 
For example, it has been shown that encoder models like BERT~\cite{devlin-etal-2019-bert}, RoBERTa~\cite{Liu2019RoBERTaAR}, and BART~\cite{lewis-etal-2020-bart} are good at pattern recognition.
Given existing insights into the impact architecture can have, and lack of insights into its impact in the context of harmful terms detection, there is value in better understanding the necessary considerations for effectively leveraging LLMs in this context.



\section{Methods}

The goal of our study is to better understand the impact of LLM architecture on the ability to detect harmful computing terms. Below we describe our methods.

\subsection{Model Selection}
\begin{figure}[H]
    \centering
    \includegraphics[width=0.9\textwidth]{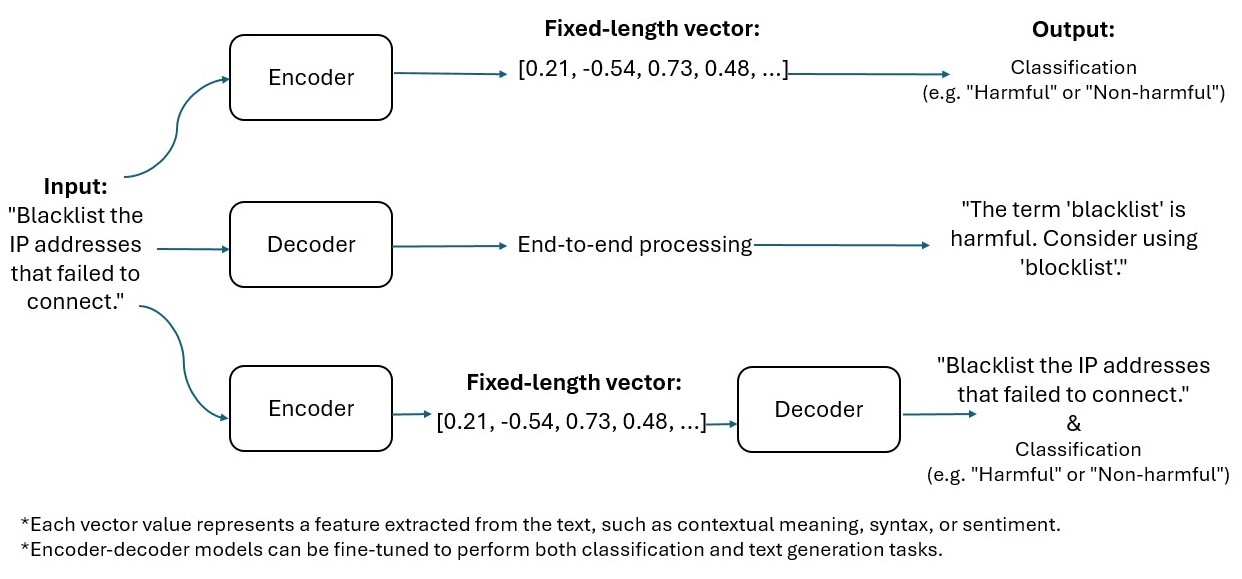}
    \caption{How Model Types Process Harmful Terms}
    \label{fig:ModelsVisual}
\end{figure}

We posit that the effectiveness of automated tools for detecting harmful and non-inclusive language depends heavily on the underlying model architecture. 
To explore the influence of different architectures on detection capabilities, we evaluated a diverse set of LLMs across architecture types (encoder, encoder-decoder, decoder). The distinction between architectures with respect to our efforts is summarized in Figure~\ref{fig:ModelsVisual}.



\paragraph{Encoder Models}
Encoder models are designed to process input text and encode it into fixed-size representations~\cite{Chadha2020DistilledEncoderDecoderEncDec}. These models excel at understanding context and semantics, making them suitable for tasks that require deep comprehension of the input text. The encoder models tested in this study include:
\begin{itemize}
    \item \textbf{BERT (Bi-directional Encoder Representations from Transformers)}: Known for its contextual understanding, BERT models can capture bi-directional relationships within the text~\cite{devlin-etal-2019-bert}. We used the \textit{bert-base-uncased} variant, which is fine-tuned for natural language inference tasks.
    \item \textbf{RoBERTa (Robustly optimized BERT approach)}: An optimized version of BERT that improves performance through better training techniques~\cite{Liu2019RoBERTaAR}. We evaluated the \textit{roberta-large-mnli} model, fine-tuned for multi-genre natural language inference.
\end{itemize}

\paragraph{Encoder-Decoder Models}
Encoder-decoder models incorporate both an encoder and a decoder~\cite{Chadha2020DistilledEncoderDecoderEncDec}. These models are adept at tasks that involve generating text based on given input, such as translation and summarization. The encoder processes the input, while the decoder generates the output. The encoder-decoder models we tested are:
\begin{itemize}
    \item \textbf{T5 (Text-To-Text Transfer Transformer)}: A versatile model that frames all NLP tasks as text-to-text transformations~\cite{10.5555/3455716.3455856}. We used the \textit{T5-large} model.
    \item \textbf{BART (Bidirectional and Auto-Regressive Transformers)}: Combines the strengths of BERT (bi-directional context) and GPT (text generation)~\cite{lewis-etal-2020-bart}. We evaluated the \textit{facebook/bart-large-mnli} model, which is fine-tuned for natural language inference.
\end{itemize}

\paragraph{Decoder Models}
Decoder models, also known as auto-regressive models, are designed to generate text sequentially~\cite{Chadha2020DistilledEncoderDecoderEncDec}. These models are particularly well-suited for tasks that require generating coherent and contextually appropriate text. The decoder models included in our study are:
\begin{itemize}
    \item \textbf{Gemini}: Versions 2.0 Flash Experimental and 1.5 Flash.~\footnote{https://gemini.google.com}
    \item \textbf{Claude AI (from Anthropic)}: Version 3.5 Sonnet.~\footnote{https://claude.ai/}
    \item \textbf{GPT-4 (ChatGPT)}: Advanced generative capabilities, particularly effective for producing human-like text.~\footnote{https://chatgpt.com}
\end{itemize}

\paragraph{Architectural Considerations}
The architectural differences between encoder, encoder-decoder, and decoder models contribute to their suitability for various tasks. For example, \textbf{encoder} models are likely best for tasks that require deep understanding and contextual analysis of the text, such as classification and sentiment analysis while \textbf{encoder-decoder} models ideal for tasks that involve both understanding the input and generating related output, such as translation and summarization. \textbf{Decoder} models may be best suited for  tasks that primarily involve text generation, such as chatbots and content creation.

\vspace{-10pt}

\paragraph{Zero-Shot Classification and Prompting}
Given the diversity of tasks and the varying strengths of different model architectures, we employed zero-shot classification techniques. Zero-shot classification enables models to handle tasks they were not explicitly trained for by providing them with task-specific prompts~\cite{HuggingFace}. This approach is particularly useful for evaluating models' abilities to detect harmful and non-inclusive language without the need for extensive task-specific fine-tuning.

To ensure meaningful outputs, we selected versions of BART, RoBERTa, and other full or partially encoder models that were fine-tuned for zero-shot classification and available through the Hugging Face zero-shot pipeline~\cite{HuggingFace}. Initial experiments using these models in their general state produced less meaningful outputs, highlighting the importance of fine-tuning for task-specific performance.

It is important to note that while encoder-decoder models like BART are traditionally designed for text generation, their flexible architecture allows them to be used for classification tasks through task-specific prompting. In our evaluation, we framed harmful term detection as a binary classification problem and prompted models to classify terms as `Harmful' or `Non-Harmful.' The decoder then generated a textual output representing the classification.

\subsection{Evaluation Setup}
Our evaluation  involved a systematic approach to evaluating LLMs for their ability to detect harmful terminology. 
This section details the prompting approach, metrics used, and the testing environment.

\subsubsection{Prompting Approach}
To ensure consistency in evaluating each model, we employed a standardized prompting approach. For each model, we provided the same prompt: 

\begin{quote}
    \textit{``Identify any harmful or non-inclusive language in the following text:''}
\end{quote}

This prompt was followed by a list of 64 harmful terms and sample use cases derived from a previously curated database of harmful technical terminology~\cite{winchester2023hate, winchester2023harmful}. 
This standardized prompt, which was consecutively given to each model three times, allowed us to assess how effectively each model could identify harmful and non-inclusive language within the given context.
Our future work will explore the impact of prompt composition and content on outcomes (Section~\ref{sec:future}).

\subsubsection{Test Environment}
The majority of our evaluations were conducted within controlled virtual environments, using Jupyter notebook (version 7.3.2) to ensure a reproducible and consistent testing setup. This approach allowed us to systematically compare the performance of various models under identical conditions. The encoder and encoder-decoder models tested each came from the Hugging Face Transformers library\footnote{https://huggingface.co/docs/transformers/v4.17.0/en/index}. For models such as ChatGPT, Claude AI, and Gemini, which are accessed through their online chat tools, we used these web-based platforms. 
These tools enabled us to harness the specific functionalities and features of each model while adhering to a standardized evaluation protocol across different platforms.

We accessed models fine-tuned for zero-shot classification, such as BART and RoBERTa, through the Hugging Face zero-shot pipeline. 
We tested each model by creating binary classification classes for ``harmful" and ``non-harmful'' and assigning each term a score representing the probability of belonging to either class. This approach leveraged the strengths of their architectures, allowing us to generate meaningful outputs without extensive task-specific fine-tuning.

\subsubsection{Metrics Used}
The primary metric used to evaluate the performance of each model was accuracy, defined as the percentage of correctly identified harmful terms out of the total 64 terms provided. 
By evaluating a range of LLMs across different architectural types, we aim to provide a comprehensive understanding of how these models perform in detecting harmful and non-inclusive terminology. This selection approach allows us to identify model-specific strengths and limitations, ultimately contributing to the development of more effective automated detection tools.

\section{Results and Analysis}

In this study, various AI models were evaluated for their ability to detect harmful or non-inclusive language. The models tested include BERT-base-uncased, RoBERTa large-mnli, T5-large, \textit{Facebook/bart-large-mnli}, Gemini Flash 2.0, Gemini Flash 1.5, Claude AI Sonnet 3.5, and GPT-4 (ChatGPT). The overall performance comparison, along with insights into each model's capabilities and limitations, provides a comprehensive understanding of their effectiveness. Results are summarized in Table~\ref{tab:results}.

\begin{table}[h]
\centering
\begin{tabular}{|l|l|c|p{3.5cm}|p{3.5cm}|}
\hline
\textbf{Model} & \textbf{Type} & \textbf{\begin{tabular}{@{}l@{}}Harmful Terms \\ Detected \end{tabular}} & \textbf{Key Strengths} & \textbf{Key Weaknesses} \\ \hline
BERT-base-uncased & Encoder & 39/64 & High accuracy, pattern recognition & Struggles with binary classification certainty, some misclassifications \\ \hline
RoBERTa large-mnli & Encoder & 11/64 & Some contextual understanding & Lower detection accuracy, misclassifies some harmful terms \\ \hline
T5-large & Enc-Dec & Inconclusive & Versatile text-to-text transformations & Repetitive, non-informative outputs in this task \\ \hline
Facebook/bart-large-mnli & Enc-Dec & 20/64 & Context awareness, higher confidence & Misclassifies some harmful terms \\ \hline
Gemini Flash 2.0 & Decoder & 44/64 & In-depth contextual analysis, high accuracy & Minor misclassifications \\ \hline
Gemini Flash 1.5 & Decoder & 42/64 & Strong context awareness, high accuracy & Minor misclassifications \\ \hline
Claude AI Sonnet 3.5 & Decoder & 37/64 & Iterative improvements, nuanced recommendations & Some variability in iterations \\ \hline
GPT-4 (ChatGPT) & Decoder & 35/64 (initially) & Nuanced recommendations, context sensitivity & Variability in detection accuracy \\ \hline
\end{tabular}
\caption{Performance Comparison of AI Models in Detecting Harmful Terms}
\label{tab:results}
\end{table}

\subsection{Overall Performance Comparison}

Among the encoder models, BERT-base-uncased demonstrated high accuracy, identifying 39 of 64 terms as harmful. The model excelled in recognizing patterns and exhibited a degree of context awareness. However, it faced challenges with binary classification, often categorizing terms with scores close to 0.50, indicating uncertainty. The model accurately identified harmful terms such as ``Blacklist" and ``Master/Slave," but misclassified terms like ``Terminate" and ``Whitelist."

RoBERTa large-mnli identified only 11 harmful terms, reflecting its lower detection accuracy in this specific context. Despite this, RoBERTa exhibited reliable performance in general language tasks due to its robust optimization, effectively processing complex language patterns. High confidence in certain classifications, such as ``Black hat" and ``Man-in-the-middle," demonstrated its contextual understanding, but the overall low detection accuracy and misclassifications highlighted its limitations.

Moving to the encoder-decoder models, T5-large provided inconclusive results, indicating its limitations in this specific task despite its versatility in text-to-text transformations. The model's outputs were generally repetitive and non-informative, failing to provide a clear classification of terms as harmful or non-harmful. This highlights the importance of task-specific fine-tuning for accurate harmful term detection~\cite{wolf-etal-2020-transformers}.

\textit{Facebook/bart-large-mnli} identified 20 harmful terms, displaying reasonable context awareness and pattern recognition. The model exhibited high confidence in classifying terms like ``Black hat," ``Blacklist," and ``Man-in-the-middle" as harmful. However, it misclassified terms such as ``Terminate," ``Whitelist," and "Grandfathered" as non-harmful despite their harmful context, indicating areas for improvement in nuanced understanding.

The decoder models, particularly the Gemini Flash models, exhibited outstanding performance. Gemini Flash 2.0 identified 44 harmful terms, making it the most accurate model in this study. The model demonstrated in-depth contextual analysis and thorough assessment, consistently identifying harmful terms across multiple prompt iterations. Similarly, Gemini Flash 1.5 identified 42 harmful terms, showing strong context awareness and accuracy. Both models used a descriptive approach rather than binary classification, reducing the likelihood of misclassifications by providing detailed context and reasoning.

Claude AI Sonnet 3.5 identified 37 harmful terms. The model showed significant improvement over iterations, providing nuanced explanations and contextual reasoning. It offered detailed suggestions and alternatives for technical terms, indicating strong contextual understanding.

GPT-4 (ChatGPT) initially identified 35 harmful terms, demonstrating excellent pattern recognition and context sensitivity. However, its detection accuracy showed variability in subsequent iterations, identifying 28 terms in the second and 24 in the third. The model's classifications became more general over time, indicating a need for refinement or other mechanisms to maintain consistent accuracy.

It is notable that some models specifically identified certain harmful terms as non-harmful. For instance, the BERT-base-uncased model classified terms like ``Average user" and ``Sanity check" as non-harmful, reflecting its limitations in this aspect. Similarly, other models like RoBERTa large-mnli and GPT-4 also exhibited this behavior, highlighting the inherent challenges in achieving consistent and accurate harmful term detection across different contexts.

\section{Discussion and Limitations}

In this section, we discuss the implications of our findings for the design of harmful term detection tools, recommendations for selecting appropriate model architectures, and the limitations and challenges encountered during the study.

\subsection{Implications for Tool Design}

The results of this study provide valuable insights for the design of tools aimed at detecting harmful or non-inclusive language. The high accuracy and detailed contextual analysis demonstrated by models like Gemini Flash 2.0 and 1.5 suggest that using models best suited for a descriptive approach, rather than binary classification, can enhance the effectiveness of such tools. These models' ability to provide nuanced explanations and in some cases automatically suggest alternative terms highlights the importance of incorporating detailed contextual reasoning into the tool design. Ultimately, tools should be designed to not only identify harmful terms but also offer explanations and constructive alternatives that promote inclusivity.

Additionally, the performance of models like Claude AI Sonnet 3.5 and BERT-base-uncased underscores the importance of combining strong pattern recognition with detailed contextual analysis. Claude AI Sonnet 3.5's iterative improvements and nuanced recommendations exemplify how a model can refine its understanding of harmful terms through repeated exposure and context analysis. On the other hand, BERT-base-uncased demonstrated high accuracy but faced confidence challenges in binary classification, suggesting a need for a more descriptive and context-aware approach.

\subsection{Architecture Selection Recommendations}

The study highlights the strengths and weaknesses of different model architectures in the context of harmful term detection. Encoder models, such as BERT-base-uncased, showed some level of accuracy but struggled with binary classification due to their tendency to categorize terms with scores close to 0.50, indicating uncertainty. These models are suitable for tasks that require strong pattern recognition and initial context awareness but may benefit from additional contextual analysis for more nuanced tasks.

Encoder-decoder models like \textit{facebook/bart-large-mnli} demonstrated reasonable context sensitivity but had lower performance compared to some decoder models. While they can handle a variety of language tasks, their performance in harmful term detection suggests a need for further refinement and task-specific fine-tuning.

Decoder models, such as Gemini Flash 2.0, Gemini Flash 1.5, and Claude AI Sonnet 3.5, exhibited exceptional performance due to their ability to provide comprehensive contextual analysis and alternative suggestions. These models are highly recommended for detecting harmful terms and promoting inclusive language. Their descriptive approach, rather than binary classification, helps reduce the likelihood of misclassifications by providing detailed context and reasoning.

Task-specific fine-tuning is crucial, as evidenced by the inconclusive results from T5-large. Despite its versatility in text-to-text transformations, T5-large failed to provide meaningful classifications in this specific task, underscoring the importance of targeted training or selection of similar models that are compatible with zero shot classification where the model does not need prior examples to generate a meaningful output.

\subsection{Limitations and Challenges}
Several limitations and challenges were encountered during the study. One significant limitation was the variability in detection accuracy across different iterations for models like GPT-4 (ChatGPT). This variability highlights the need for consistent performance and refinement to ensure reliable results. Additionally, some models, such as RoBERTa large-mnli, demonstrated lower detection accuracy, indicating the challenges in achieving consistent harmful term detection across various contexts. The study also revealed the limitations of models best suited for binary classification approaches, which often resulted in misclassifications. Implementing a descriptive approach with detailed contextual analysis, best performed by fine-tuned decoder models, can mitigate these challenges.

\section{Key Findings \& Future Work}\label{sec:future}
Our study highlights the critical role of model architecture in using LLMs to detect harmful and non-inclusive terminology in technical contexts. Decoder models like Gemini Flash and Claude AI demonstrate exceptional contextual understanding and the ability to provide nuanced suggestions, while encoder models such as BERT exhibit strong pattern recognition but face limitations in binary classification tasks. Our findings emphasize the importance of leveraging descriptive approaches over binary classifications to improve detection accuracy and inclusivity. Future research should focus on refining task-specific prompts, integrating user feedback mechanisms, and expanding datasets to enhance the adaptability of inclusive language tools across diverse domains. By advancing these tools, we aim to contribute to a more equitable and welcoming technical community.

Our next steps are to evaluate whether these tools are able to better understand context, provide helpful replacement suggestions, and handle more complex scenarios. For example, tools could ask users to replace harmful words in a sentence with prompts such as `Replace noninclusive terms in the following sentence with more inclusive alternatives: [We had to abort the project due to the lack of funding].' or suggest alternatives for a list of terms with prompts such as `Suggest inclusive alternatives for harmful terms in this list: [List: master, minority, blacklist].' These steps will help make the tools more accurate and thoughtful in the suggestions they provide.  Another goal is to improve tools so that they can explain why a term is harmful and offer better replacements based on the situation, such as with prompts like `Explain why the term `[brown bags]' is considered harmful and provide a suitable replacement.' For more complicated tasks, such as editing technical documents or rewriting code comments, tools could use prompts like, “Analyze the following technical document and replace any harmful terminology with inclusive language: [The master node controls the slave nodes in the system. If a node is behaving abnormally, it is added to the blacklist and prevented from further communication”, or “Rewrite this code comment to remove harmful language and ensure it is inclusive: [If a thread fails, it is added to the blacklist to prevent further execution].” These improvements will help inclusive language tools work better in different situations and make them more useful for everyone.


\bibliographystyle{ACM-Reference-Format}
\bibliography{references}


\begin{thebibliography}{19}


\ifx \showCODEN    \undefined \def \showCODEN     #1{\unskip}     \fi
\ifx \showISBNx    \undefined \def \showISBNx     #1{\unskip}     \fi
\ifx \showISBNxiii \undefined \def \showISBNxiii  #1{\unskip}     \fi
\ifx \showISSN     \undefined \def \showISSN      #1{\unskip}     \fi
\ifx \showLCCN     \undefined \def \showLCCN      #1{\unskip}     \fi
\ifx \shownote     \undefined \def \shownote      #1{#1}          \fi
\ifx \showarticletitle \undefined \def \showarticletitle #1{#1}   \fi
\ifx \showURL      \undefined \def \showURL       {\relax}        \fi
\providecommand\bibfield[2]{#2}
\providecommand\bibinfo[2]{#2}
\providecommand\natexlab[1]{#1}
\providecommand\showeprint[2][]{arXiv:#2}

\bibitem[Chadha(2020)]%
        {Chadha2020DistilledEncoderDecoderEncDec}
\bibfield{author}{\bibinfo{person}{Aman Chadha}.} \bibinfo{year}{2020}\natexlab{}.
\newblock \showarticletitle{Encoder vs. Decoder vs. Encoder-Decoder Models}.
\newblock \bibinfo{journal}{\emph{Distilled AI}} (\bibinfo{year}{2020}).
\newblock
\newblock
\shownote{\url{https://aman.ai}}.


\bibitem[Devlin et~al\mbox{.}(2019)]%
        {devlin-etal-2019-bert}
\bibfield{author}{\bibinfo{person}{Jacob Devlin}, \bibinfo{person}{Ming-Wei Chang}, \bibinfo{person}{Kenton Lee}, {and} \bibinfo{person}{Kristina Toutanova}.} \bibinfo{year}{2019}\natexlab{}.
\newblock \showarticletitle{{BERT}: Pre-training of Deep Bidirectional Transformers for Language Understanding}. In \bibinfo{booktitle}{\emph{Proceedings of the 2019 Conference of the North {A}merican Chapter of the Association for Computational Linguistics: Human Language Technologies, Volume 1 (Long and Short Papers)}}, \bibfield{editor}{\bibinfo{person}{Jill Burstein}, \bibinfo{person}{Christy Doran}, {and} \bibinfo{person}{Thamar Solorio}} (Eds.). \bibinfo{publisher}{Association for Computational Linguistics}, \bibinfo{address}{Minneapolis, Minnesota}.
\newblock
\href{https://doi.org/10.18653/v1/N19-1423}{doi:\nolinkurl{10.18653/v1/N19-1423}}


\bibitem[Diversity and Council(2022)]%
        {words2022acm}
\bibfield{author}{\bibinfo{person}{ACM Diversity} {and} \bibinfo{person}{Inclusion Council}.} \bibinfo{year}{2022}\natexlab{}.
\newblock \bibinfo{title}{Words Matter: Alternatives for Charged Terminology in the Computing Profession}.
\newblock
\urldef\tempurl%
\url{https://www.acm.org/diversity-inclusion/words-matter}
\showURL{%
\tempurl}


\bibitem[Face({[n.\,d.]})]%
        {HuggingFace}
\bibfield{author}{\bibinfo{person}{Hugging Face}.} \bibinfo{year}{[n.\,d.]}\natexlab{}.
\newblock \bibinfo{title}{What is Zero-Shot Classification? - Hugging Face}.
\newblock
\urldef\tempurl%
\url{https://huggingface.co/tasks/zero-shot-classification}
\showURL{%
\tempurl}


\bibitem[Giachanou and Rosso(2020)]%
        {giachanou2020battle}
\bibfield{author}{\bibinfo{person}{Anastasia Giachanou} {and} \bibinfo{person}{Paolo Rosso}.} \bibinfo{year}{2020}\natexlab{}.
\newblock \showarticletitle{The Battle Against Online Harmful Information: The Cases of Fake News and Hate Speech}. \bibinfo{publisher}{Association for Computing Machinery}, \bibinfo{address}{New York, NY, USA}.
\newblock
\showISBNx{9781450368599}
\href{https://doi.org/10.1145/3340531.3412169}{doi:\nolinkurl{10.1145/3340531.3412169}}


\bibitem[Gilbert et~al\mbox{.}(2022)]%
        {gilbert2022words}
\bibfield{author}{\bibinfo{person}{Juan~E Gilbert}, \bibinfo{person}{Stephanie Ludi}, \bibinfo{person}{David~A Patterson}, {and} \bibinfo{person}{Lisa~M Smith}.} \bibinfo{year}{2022}\natexlab{}.
\newblock \showarticletitle{Words matter}.
\newblock \bibinfo{journal}{\emph{Commun. ACM}} \bibinfo{volume}{65}, \bibinfo{number}{7} (\bibinfo{year}{2022}), \bibinfo{pages}{36--36}.
\newblock


\bibitem[Kumar et~al\mbox{.}(2024)]%
        {kumar2024investigating}
\bibfield{author}{\bibinfo{person}{Divyanshu Kumar}, \bibinfo{person}{Umang Jain}, \bibinfo{person}{Sahil Agarwal}, {and} \bibinfo{person}{Prashanth Harshangi}.} \bibinfo{year}{2024}\natexlab{}.
\newblock \showarticletitle{Investigating Implicit Bias in Large Language Models: A Large-Scale Study of Over 50 LLMs}.
\newblock \bibinfo{journal}{\emph{arXiv preprint arXiv:2410.12864}} (\bibinfo{year}{2024}).
\newblock


\bibitem[Lewis et~al\mbox{.}(2020)]%
        {lewis-etal-2020-bart}
\bibfield{author}{\bibinfo{person}{Mike Lewis}, \bibinfo{person}{Yinhan Liu}, \bibinfo{person}{Naman Goyal}, \bibinfo{person}{Marjan Ghazvininejad}, \bibinfo{person}{Abdelrahman Mohamed}, \bibinfo{person}{Omer Levy}, \bibinfo{person}{Veselin Stoyanov}, {and} \bibinfo{person}{Luke Zettlemoyer}.} \bibinfo{year}{2020}\natexlab{}.
\newblock \showarticletitle{{BART}: Denoising Sequence-to-Sequence Pre-training for Natural Language Generation, Translation, and Comprehension}. In \bibinfo{booktitle}{\emph{Proceedings of the 58th Annual Meeting of the Association for Computational Linguistics}}, \bibfield{editor}{\bibinfo{person}{Dan Jurafsky}, \bibinfo{person}{Joyce Chai}, \bibinfo{person}{Natalie Schluter}, {and} \bibinfo{person}{Joel Tetreault}} (Eds.). \bibinfo{publisher}{Association for Computational Linguistics}, \bibinfo{address}{Online}.
\newblock
\href{https://doi.org/10.18653/v1/2020.acl-main.703}{doi:\nolinkurl{10.18653/v1/2020.acl-main.703}}


\bibitem[Liu et~al\mbox{.}(2019)]%
        {Liu2019RoBERTaAR}
\bibfield{author}{\bibinfo{person}{Yinhan Liu}, \bibinfo{person}{Myle Ott}, \bibinfo{person}{Naman Goyal}, \bibinfo{person}{Jingfei Du}, \bibinfo{person}{Mandar Joshi}, \bibinfo{person}{Danqi Chen}, \bibinfo{person}{Omer Levy}, \bibinfo{person}{Mike Lewis}, \bibinfo{person}{Luke Zettlemoyer}, {and} \bibinfo{person}{Veselin Stoyanov}.} \bibinfo{year}{2019}\natexlab{}.
\newblock \showarticletitle{RoBERTa: A Robustly Optimized BERT Pretraining Approach}.
\newblock \bibinfo{journal}{\emph{ArXiv}}  \bibinfo{volume}{abs/1907.11692} (\bibinfo{year}{2019}).
\newblock
\urldef\tempurl%
\url{https://api.semanticscholar.org/CorpusID:198953378}
\showURL{%
\tempurl}


\bibitem[May et~al\mbox{.}(2019)]%
        {may-etal-2019-measuring}
\bibfield{author}{\bibinfo{person}{Chandler May}, \bibinfo{person}{Alex Wang}, \bibinfo{person}{Shikha Bordia}, \bibinfo{person}{Samuel~R. Bowman}, {and} \bibinfo{person}{Rachel Rudinger}.} \bibinfo{year}{2019}\natexlab{}.
\newblock \showarticletitle{On Measuring Social Biases in Sentence Encoders}. In \bibinfo{booktitle}{\emph{Proceedings of the 2019 Conference of the North {A}merican Chapter of the Association for Computational Linguistics: Human Language Technologies, Volume 1 (Long and Short Papers)}}, \bibfield{editor}{\bibinfo{person}{Jill Burstein}, \bibinfo{person}{Christy Doran}, {and} \bibinfo{person}{Thamar Solorio}} (Eds.). \bibinfo{publisher}{Association for Computational Linguistics}, \bibinfo{address}{Minneapolis, Minnesota}, \bibinfo{pages}{622--628}.
\newblock
\href{https://doi.org/10.18653/v1/N19-1063}{doi:\nolinkurl{10.18653/v1/N19-1063}}


\bibitem[Raffel et~al\mbox{.}(2020)]%
        {10.5555/3455716.3455856}
\bibfield{author}{\bibinfo{person}{Colin Raffel}, \bibinfo{person}{Noam Shazeer}, \bibinfo{person}{Adam Roberts}, \bibinfo{person}{Katherine Lee}, \bibinfo{person}{Sharan Narang}, \bibinfo{person}{Michael Matena}, \bibinfo{person}{Yanqi Zhou}, \bibinfo{person}{Wei Li}, {and} \bibinfo{person}{Peter~J. Liu}.} \bibinfo{year}{2020}\natexlab{}.
\newblock \showarticletitle{Exploring the limits of transfer learning with a unified text-to-text transformer}.
\newblock \bibinfo{journal}{\emph{J. Mach. Learn. Res.}} \bibinfo{volume}{21}, \bibinfo{number}{1}, Article \bibinfo{articleno}{140} (\bibinfo{date}{Jan.} \bibinfo{year}{2020}).
\newblock
\showISSN{1532-4435}


\bibitem[Todd et~al\mbox{.}(2024)]%
        {todd2024githubinclusifier}
\bibfield{author}{\bibinfo{person}{Liam Todd}, \bibinfo{person}{John Grundy}, {and} \bibinfo{person}{Christoph Treude}.} \bibinfo{year}{2024}\natexlab{}.
\newblock \showarticletitle{GitHubInclusifier: Finding and fixing non-inclusive language in GitHub Repositories}. In \bibinfo{booktitle}{\emph{Proceedings of the 2024 IEEE/ACM 46th International Conference on Software Engineering: Companion Proceedings}}. \bibinfo{pages}{89--93}.
\newblock


\bibitem[Vanderborght and Okamura(2020)]%
        {vanderborght2020united}
\bibfield{author}{\bibinfo{person}{Bram Vanderborght} {and} \bibinfo{person}{Allison Okamura}.} \bibinfo{year}{2020}\natexlab{}.
\newblock \showarticletitle{United against racism and a call for action [Ethical, legal, and societal issues]}.
\newblock \bibinfo{journal}{\emph{IEEE Robotics \& Automation Magazine}} \bibinfo{volume}{27}, \bibinfo{number}{3} (\bibinfo{year}{2020}), \bibinfo{pages}{10--11}.
\newblock


\bibitem[Varona and Su{\'a}rez(2022)]%
        {varona2022discrimination}
\bibfield{author}{\bibinfo{person}{Daniel Varona} {and} \bibinfo{person}{Juan~Luis Su{\'a}rez}.} \bibinfo{year}{2022}\natexlab{}.
\newblock \showarticletitle{Discrimination, bias, fairness, and trustworthy AI}.
\newblock \bibinfo{journal}{\emph{Applied Sciences}} \bibinfo{volume}{12}, \bibinfo{number}{12} (\bibinfo{year}{2022}), \bibinfo{pages}{5826}.
\newblock


\bibitem[Warr et~al\mbox{.}(2024)]%
        {warr2024implicit}
\bibfield{author}{\bibinfo{person}{Melissa Warr}, \bibinfo{person}{Nicole~Jakubczyk Oster}, {and} \bibinfo{person}{Roger Isaac}.} \bibinfo{year}{2024}\natexlab{}.
\newblock \showarticletitle{Implicit bias in large language models: Experimental proof and implications for education}.
\newblock \bibinfo{journal}{\emph{Journal of Research on Technology in Education}} (\bibinfo{year}{2024}), \bibinfo{pages}{1--24}.
\newblock


\bibitem[Winchester et~al\mbox{.}(2023a)]%
        {winchester2023hate}
\bibfield{author}{\bibinfo{person}{Hana Winchester}, \bibinfo{person}{Ebtesam Al~Haque}, \bibinfo{person}{Alicia Boyd}, {and} \bibinfo{person}{Brittany Johnson}.} \bibinfo{year}{2023}\natexlab{a}.
\newblock \showarticletitle{HaTe Detector: A Tool for Detecting and Correcting Harmful Terminology in Computing Artifacts}. In \bibinfo{booktitle}{\emph{2023 IEEE Symposium on Visual Languages and Human-Centric Computing (VL/HCC)}}. IEEE, \bibinfo{pages}{245--248}.
\newblock


\bibitem[Winchester et~al\mbox{.}(2022)]%
        {winchester2022exploration}
\bibfield{author}{\bibinfo{person}{Hana Winchester}, \bibinfo{person}{Alicia~E Boyd}, {and} \bibinfo{person}{Brittany Johnson}.} \bibinfo{year}{2022}\natexlab{}.
\newblock \showarticletitle{An Exploration of Intersectionality in Software Development and Use}. In \bibinfo{booktitle}{\emph{2022 IEEE/ACM 3rd International Workshop on Gender Equality, Diversity and Inclusion in Software Engineering (GEICSE)}}. IEEE, \bibinfo{pages}{67--70}.
\newblock


\bibitem[Winchester et~al\mbox{.}(2023b)]%
        {winchester2023harmful}
\bibfield{author}{\bibinfo{person}{Hana Winchester}, \bibinfo{person}{Alicia~E Boyd}, {and} \bibinfo{person}{Brittany Johnson}.} \bibinfo{year}{2023}\natexlab{b}.
\newblock \showarticletitle{Harmful Terms in Computing: Towards Widespread Detection and Correction}. In \bibinfo{booktitle}{\emph{2023 IEEE/ACM 45th International Conference on Software Engineering: Software Engineering in Society (ICSE-SEIS)}}. IEEE, \bibinfo{pages}{132--137}.
\newblock


\bibitem[Wolf et~al\mbox{.}(2020)]%
        {wolf-etal-2020-transformers}
\bibfield{author}{\bibinfo{person}{Thomas Wolf}, \bibinfo{person}{Lysandre Debut}, \bibinfo{person}{Victor Sanh}, \bibinfo{person}{Julien Chaumond}, \bibinfo{person}{Clement Delangue}, \bibinfo{person}{Anthony Moi}, \bibinfo{person}{Pierric Cistac}, \bibinfo{person}{Tim Rault}, \bibinfo{person}{Remi Louf}, \bibinfo{person}{Morgan Funtowicz}, \bibinfo{person}{Joe Davison}, \bibinfo{person}{Sam Shleifer}, \bibinfo{person}{Patrick von Platen}, \bibinfo{person}{Clara Ma}, \bibinfo{person}{Yacine Jernite}, \bibinfo{person}{Julien Plu}, \bibinfo{person}{Canwen Xu}, \bibinfo{person}{Teven Le~Scao}, \bibinfo{person}{Sylvain Gugger}, \bibinfo{person}{Mariama Drame}, \bibinfo{person}{Quentin Lhoest}, {and} \bibinfo{person}{Alexander Rush}.} \bibinfo{year}{2020}\natexlab{}.
\newblock \showarticletitle{Transformers: State-of-the-Art Natural Language Processing}. In \bibinfo{booktitle}{\emph{Proceedings of the 2020 Conference on Empirical Methods in Natural Language Processing: System Demonstrations}}, \bibfield{editor}{\bibinfo{person}{Qun Liu} {and} \bibinfo{person}{David Schlangen}} (Eds.). \bibinfo{publisher}{Association for Computational Linguistics}, \bibinfo{address}{Online}, \bibinfo{pages}{38--45}.
\newblock
\href{https://doi.org/10.18653/v1/2020.emnlp-demos.6}{doi:\nolinkurl{10.18653/v1/2020.emnlp-demos.6}}


\end{thebibliography}


\end{document}